\documentclass[11pt,letterpaper]{article}

\usepackage[T1]{fontenc}
\usepackage{newtxtext,newtxmath}
\usepackage[margin=1in]{geometry}
\usepackage{microtype}
\usepackage[hyphens]{url}
\usepackage[hidelinks]{hyperref}
\usepackage{graphicx}
\usepackage{caption}
\usepackage{booktabs}
\usepackage{array}
\usepackage{tabularx}
\usepackage[section]{placeins}
\usepackage{amsmath}
\usepackage{pifont}
\usepackage{algorithm}
\usepackage{algorithmic}
\usepackage{natbib}
\usepackage{authblk}

\graphicspath{{Figures/}}
\title{ExceptionDrive: A Planning-Oriented Counterfactual Corner-Case Benchmark for Autonomous Driving}

\author[1]{Ziyi Luo}
\author[1]{Zhe Sun}
\author[1]{Yehao Lu}
\author[2]{Lei Zhou}
\author[2]{Lisheng Wu}
\author[1]{Xuewei Li}
\author[1]{Zequn Qin}
\author[1]{Xi Li\thanks{Corresponding author: \href{mailto:xilizju@zju.edu.cn}{xilizju@zju.edu.cn}.}}
\affil[1]{College of Computer Science and Technology, Zhejiang University, Hangzhou, China}
\affil[2]{Yinwang Intelligent Technology Co., Ltd., Shenzhen, China}
\date{}

\begin{document}
\maketitle

\begin{abstract}

Average performance on routine driving benchmarks does not establish planner reliability under rare, safety-critical hazards. We proposed ExceptionDrive, a counterfactual planning benchmark that uses VLM-assisted screening, localized multi-view editing, and quality auditing to insert hazards into real nuScenes scenes while preserving their context. Its 21 tasks span six safety families and define hazard or conflict regions, local safety constraints, and acceptable responses. Because hazard insertion can invalidate the recorded human trajectory, our reference-free protocol evaluates edited predictions using Unsafe Rate (UR), Hazard Clearance Compliance (HCC), Hazard Proximity Response (HPR), and Counterfactual Trajectory Shift (CTS), which measure core-region intrusion, clearance compliance, clearance relative to a prescribed margin, and counterfactual trajectory change. Seven representative planners frequently intrude into hazard regions or provide insufficient clearance. We also develop a Reminder Agent that, without sample-specific task labels, converts visual evidence and the shared taxonomy into structured records of hazard presence, type, and a recommended high-level strategy. The agent neither predicts trajectories nor controls the vehicle; its records guide a VLM-based decision agent. In zero-shot experiments, the reminders improve strategy accuracy and reduce under-warning.

\end{abstract}

\begin{figure*}[!t]
    \centering
    \includegraphics[width=0.95\textwidth]{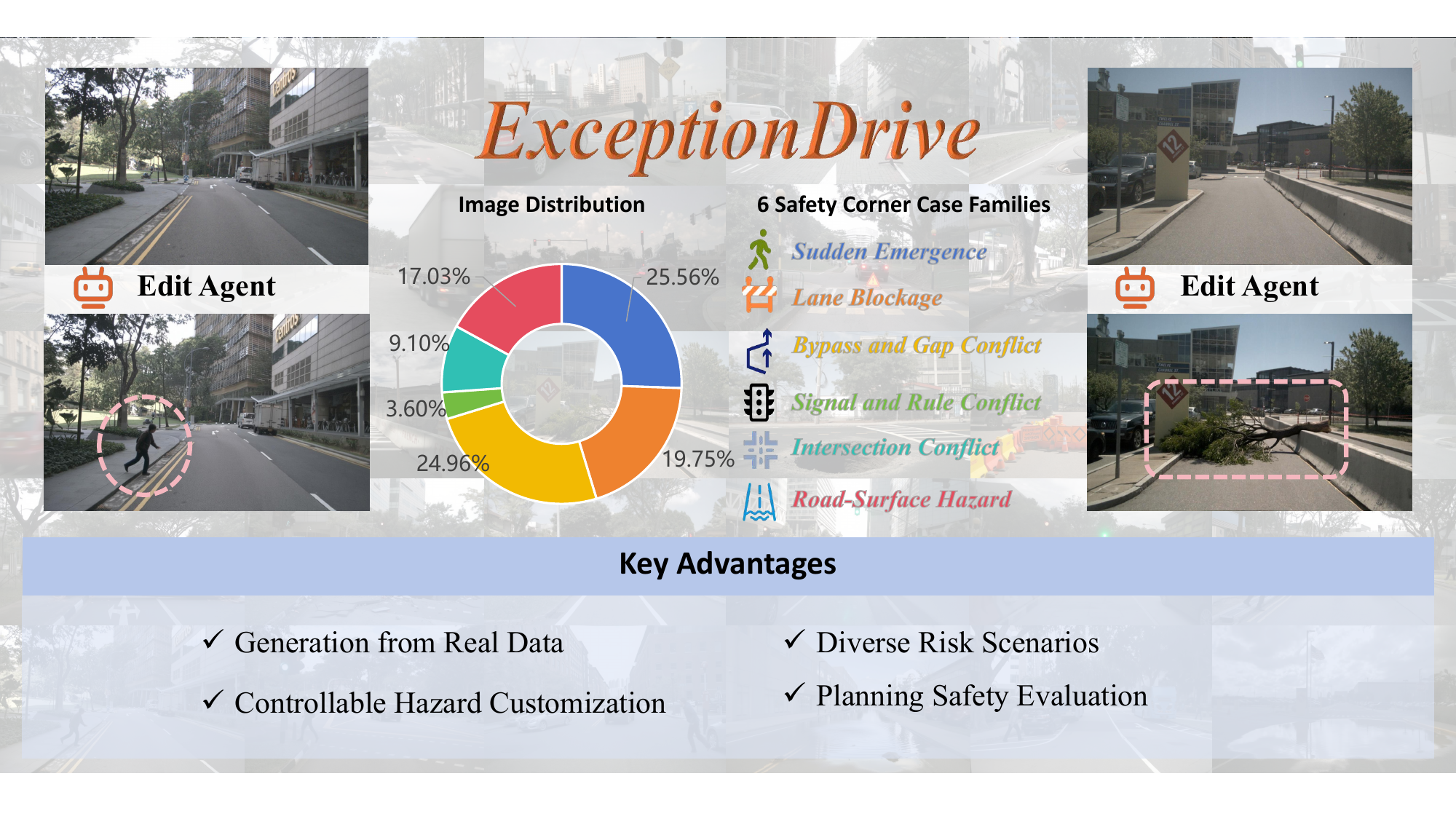}
    \caption{Overview of ExceptionDrive. Our framework generates localized counterfactual hazards in real driving scenes, evaluates planner responses without requiring an expert reference trajectory for the edited scene, and produces structured safety-aware reminders. The benchmark includes 21 planning tasks across six planning-safety families.}
    \label{first_img}
\end{figure*}

\section{Introduction}

Autonomous-driving planners have achieved strong performance on standard benchmarks, yet these benchmarks remain dominated by routine, low-risk scenes. Average-case performance therefore does not establish reliability under rare, safety-critical hazards, such as an occluded pedestrian entering the ego path or a collapsed road surface blocking the lane.

Existing resources provide limited support for controlled planner evaluation in such cases. Real-world datasets preserve realistic observations but contain few controllable hazards, whereas simulation introduces a domain gap. Corner-case datasets often target perception, anomaly recognition, or semantic reasoning rather than trajectory compliance. A planning benchmark should instead combine realistic scenes, controlled hazard insertion, paired source--edited observations, spatial hazard annotations, and trajectory-level evaluation.

Hazard insertion also invalidates conventional trajectory supervision: the recorded human trajectory may intersect the new hazard and no longer provide a valid reference. Moreover, slowing, stopping, yielding, and bypassing may all be acceptable under different constraints. Each task should therefore define local safety constraints and a set of acceptable responses rather than prescribe a unique expert trajectory.

We proposed \textbf{ExceptionDrive}, a planning-oriented counterfactual benchmark built by inserting localized hazards into real nuScenes observations. Its construction pipeline combines VLM-assisted scene screening, localized multi-view editing, quality review, and BEV hazard localization while preserving the surrounding context. The benchmark contains \textbf{21 tasks in six safety families}. Overall, ExceptionDrive comprises \textbf{12,440 source--edited pairs}: 3,181 quality-approved benchmark pairs and 9,259 auxiliary pairs. Source--edited pairs use identical model settings and non-visual inputs, allowing the effect of the localized edit to be isolated.

Our hazard-centric protocol evaluates edited trajectories without expert references. \textbf{Unsafe Rate (UR)} measures core-region intrusion, and \textbf{Hazard Clearance Compliance (HCC)} measures compliance with the required clearance. \textbf{Hazard Proximity Response (HPR)} diagnoses clearance relative to the prescribed margin, while \textbf{Counterfactual Trajectory Shift (CTS)} measures the change between source and edited predictions. HCC and UR are the primary safety metrics; HPR and CTS are complementary diagnostics. The protocol evaluates local hazard compliance rather than complete route validity or closed-loop safety.

Experiments on representative planners reveal frequent intrusion and insufficient clearance. We also develop a lightweight \textbf{Reminder Agent} that maps visual evidence and the shared task taxonomy to a structured record containing hazard status, type, relative location, risk level, supporting evidence, the applicable safety constraint, and a recommended high-level strategy. Without receiving sample-specific task labels or target strategies, the reminder supports high-level decision making but does not predict trajectories or control the vehicle.

Our main contributions are threefold:
\begin{enumerate}
    \item We proposed \textbf{ExceptionDrive}, a planning-oriented counterfactual corner-case benchmark created through localized  multi-view editing of real driving scenes. It enables controlled insertion of safety-critical hazards while preserving the surrounding driving context as much as possible.

    \item We established a planning-oriented task taxonomy and hazard-centric evaluation framework. The framework organizes 21 tasks into six planning-safety families, defines each task through local safety constraints and acceptable risk-mitigation responses, and evaluates edited-scene trajectories without requiring expert reference trajectories. HCC, UR, HPR, and CTS respectively characterize clearance compliance, core-region intrusion, hazard-proximity response, and counterfactual trajectory change.

    \item We systematically evaluated representative planners using the proposed benchmark and reveal frequent failures in local hazard compliance. As an auxiliary application, we also introduced a lightweight Reminder Agent and found that structured task-grounded risk records improve high-level strategy accuracy and reduce under-warning in zero-shot VLM experiments.
\end{enumerate}

\section{Related Work}
\subsection{VLM-Assisted Data Generation for Autonomous Driving}

Autonomous-driving data come from real-world logs, simulation, and generative models. Real-world datasets preserve authentic distributions but are costly and dominated by routine driving~\cite{nuscenes,argoverse,waymo,kitti}; simulation offers controllability but introduces domain gaps~\cite{carla,bench2drive,sdac}; and generative models synthesize scenes conditioned on text, layout, or geometry~\cite{magicdrive,panacea,drivingdiffusion,drivedreamer}.

VLMs support scene captioning, semantic annotation, prompt generation, and quality assessment~\cite{drivelm,codaLM,omnidrive,styledrive}. Rather than synthesizing entire scenes, ExceptionDrive uses VLMs to select task-compatible real scenes, generate location-aware editing instructions, and review edited samples. This localized approach retains the source scene as a within-scene control while seeking to isolate the intended hazard.
\subsection{Corner-Case Taxonomies and Dataset Construction}

Corner cases span multiple levels, from sensor or appearance shifts and unusual objects to rare scene configurations and safety-critical interactions~\cite{systematization,application_corner,space_time}. Existing benchmarks can be broadly grouped by their evaluation targets. Perception-oriented datasets assess road-anomaly segmentation or unknown-object detection~\cite{lostandfound,fishyscapes,segmentmeifyoucan,coda,sdac}; event- and reasoning-oriented datasets collect accidents or annotate rare scenes with semantic descriptions and driving suggestions~\cite{dada,codaLM}; and system-level benchmarks evaluate driving policies in interactive simulation~\cite{carla,longest6,safebench,bench2drive}.

Real-world datasets preserve visual realism but provide limited control over rare hazards, whereas simulation offers controllability at the cost of a domain gap. ExceptionDrive complements these settings with planner-level counterfactual testing on paired real scenes.

\section{Method}
\begin{figure}[!htbp]
    \centering
    \includegraphics[width=0.95\textwidth]{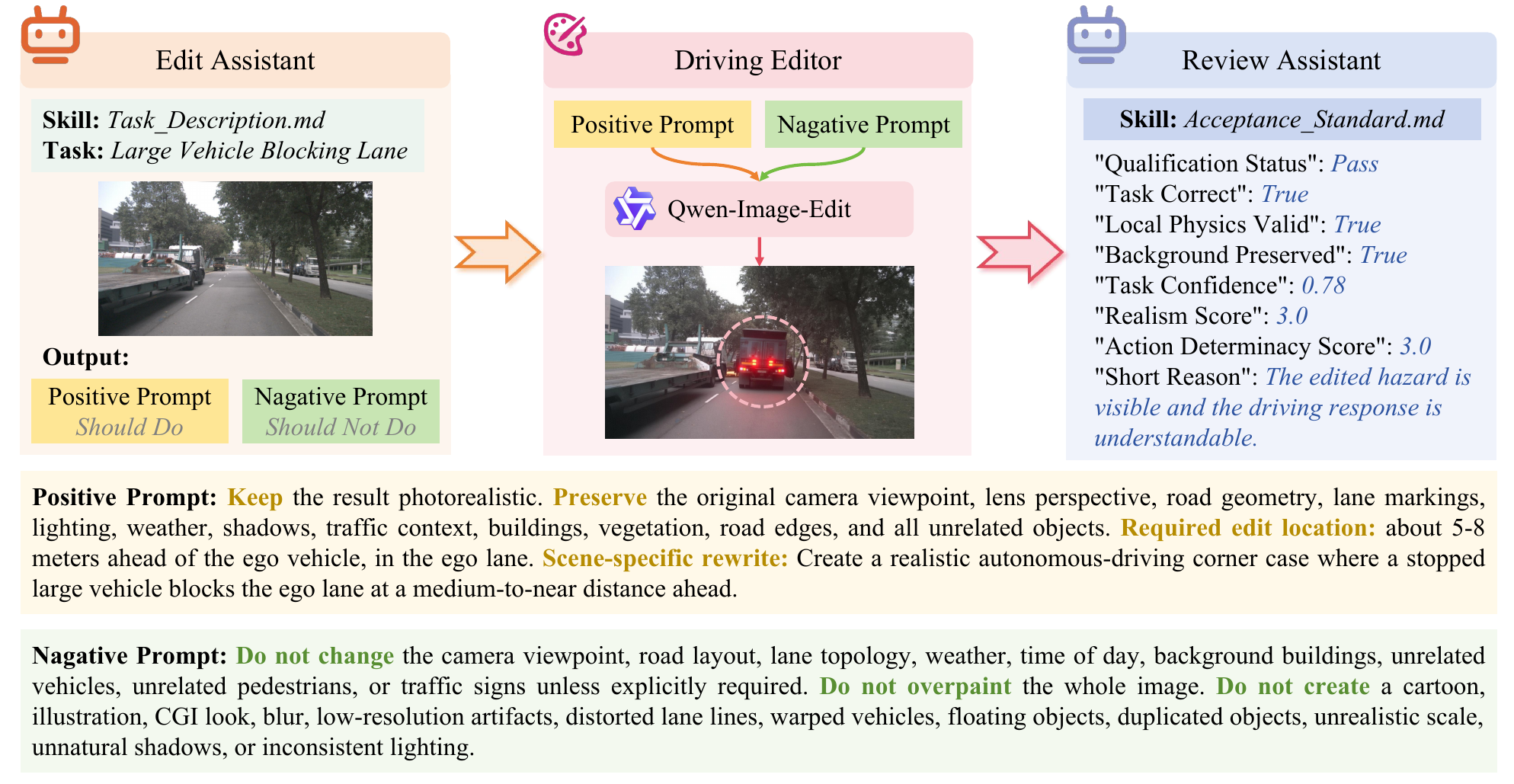}
    \caption{Counterfactual corner-case construction pipeline. An edit assistant uses the task definition and real-scene context to produce task-aware positive and negative prompts. A localized driving-scene editor inserts the target hazard while preserving viewpoint, road geometry, lighting, traffic context, and unrelated background. A review assistant then audits task correctness, physical plausibility, background preservation, realism, and action determinacy. Accepted source--edit pairs provide hazard-localized, planning-oriented samples for evaluation.}
\end{figure}

\subsection{Task Taxonomy}
\label{ssec:Taxonomy}

An object- or appearance-based corner-case taxonomy is insufficient for planner evaluation because risk depends on the hazard's location, its conflict with the ego vehicle, and the feasible responses. For example, the same vehicle may require stopping when it blocks the ego lane, yielding when it occupies an intersection conflict region, or performing a constrained bypass when sufficient adjacent space is available. Conversely, visually different hazards may impose the same planning constraint, such as avoiding a non-drivable region. Constructing planning-oriented counterfactual cases therefore requires an operational task specification that jointly defines the applicable driving context, the localized hazard intervention, the spatial safety constraint, and the range of acceptable risk-mitigation responses.

We define 21 tasks and organize them into six planning-safety families: \emph{Sudden Emergence}, \emph{Lane Blockage}, \emph{Bypass and Gap Conflict}, \emph{Signal and Rule Conflict}, \emph{Intersection Conflict}, and \emph{Road-Surface Hazard}. For each task, we specify scene-suitability conditions, rejection rules, edit content, target-location constraints, the corresponding hazard or conflict region, and acceptable risk-mitigation responses. These responses do not prescribe a single mandatory maneuver. Instead, they describe task-level options such as avoiding a localized risk region, stopping before a constraint boundary, yielding to conflicting traffic, or bypassing only when sufficient space is available. The resulting specifications provide a common basis for scene screening, localized editing, quality review, hazard-region construction, geometric planner evaluation, and high-level strategy evaluation.

\subsection{Dataset Construction}
\label{ssec:Construction}

ExceptionDrive uses task-conditioned local counterfactual editing to introduce planning-relevant hazards while preserving the surrounding real-world context. The task specifications in Section~\ref{ssec:Taxonomy} guide four stages: scene screening, localized multi-view editing, quality review, and hazard-region localization.

For each scene--task combination, a VLM assesses whether the synchronized nuScenes views and scene description satisfy the task-specific suitability conditions. It examines the road layout, available space, occlusion pattern, and potential insertion location. For suitable combinations, the VLM selects the relevant views, specifies an ego-relative target location, and generates view-specific editing instructions. A combination is rejected if the hazard would violate task constraints, create an ambiguous planning conflict, or require substantial changes to the original scene.

An instruction-driven editor then combines global preservation requirements with the task-level hazard description and scene-specific location instructions. It modifies only the target region while preserving the viewpoint, road geometry, lane markings, illumination, weather, unrelated traffic participants, and background. For hazards visible in multiple synchronized views, view-specific instructions promote cross-view coherence, which is subsequently assessed during VLM review; no separate geometric multi-view consistency module is used. The resulting source--edited pair is intended to differ primarily in the inserted hazard.

Each edit undergoes task-aware quality review. A VLM reviewer assesses hazard correctness, physical plausibility, visibility, cross-view consistency, source-scene preservation, and clarity of the planning conflict. Low-scoring edits and a random subset of accepted edits are also inspected manually. Only pairs satisfying both visual-quality and task-validity requirements are retained.

Finally, we construct a hazard region \(H_i\) in ego coordinates for each accepted sample. BEVFormer-based localization~\cite{bevformer} provides four ground-plane corners defining \(H_i\) as a closed BEV quadrilateral. For physical hazards, \(H_i\) represents the occupied or non-drivable region; for signal, right-of-way, and intersection tasks, it represents the task-defined conflict or no-go region. Each region is verified against the intended edit location, multi-view evidence, camera calibration, and task definition, and serves as the spatial reference for hazard-centric evaluation.

\subsection{Paired Counterfactual Planner Evaluation}
\label{ssec:Evaluation}

\begin{figure}[!htbp]
    \centering
    \includegraphics[width=0.78\linewidth]{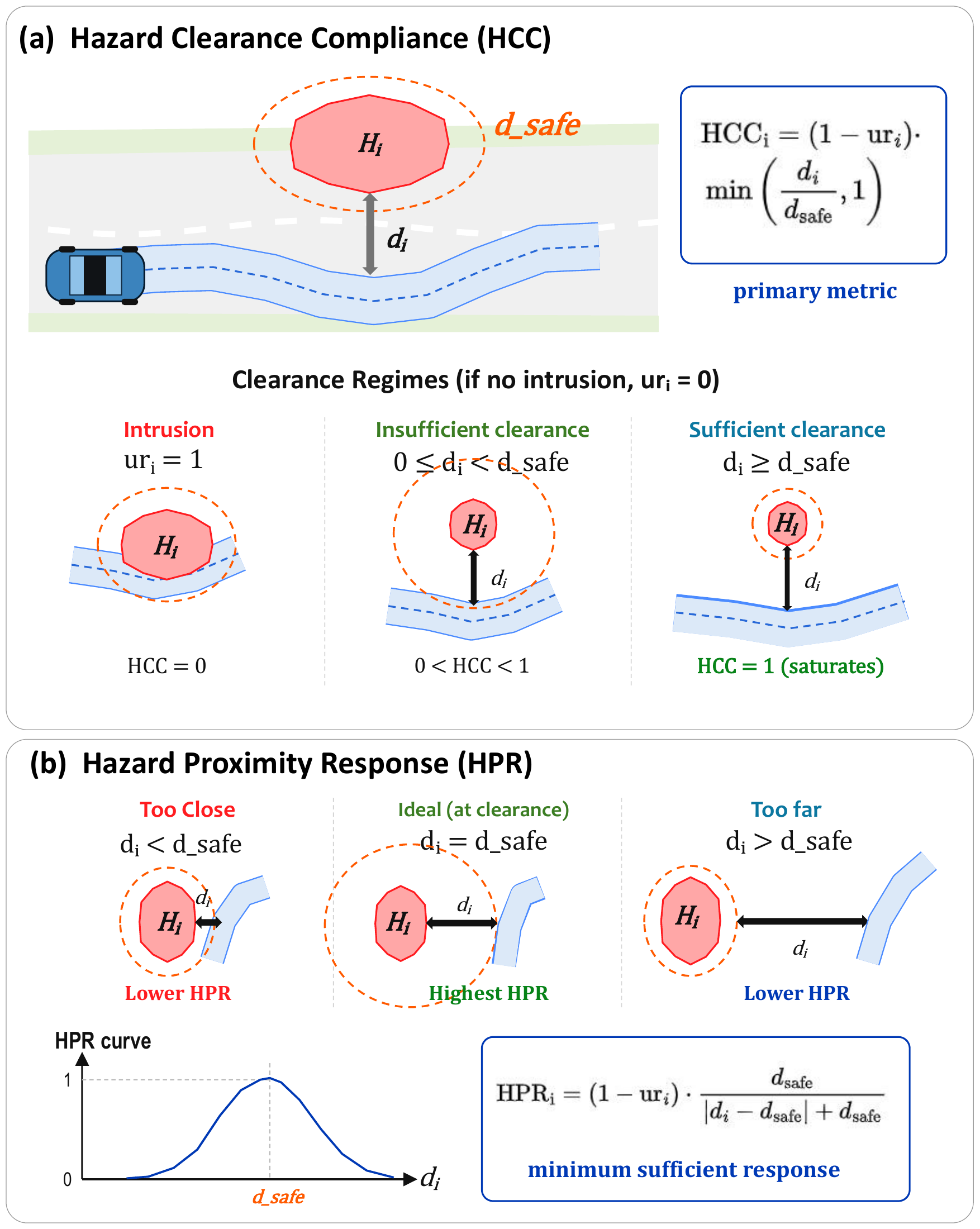}
    \captionsetup{font=small,skip=3pt}
    \caption{Illustration of the core hazard-centric evaluation metrics used in ExceptionDrive. The blue collision tube is evaluated relative to the task-defined hazard $H_i$ and its required clearance boundary $d_{\mathrm{safe}}$. \textbf{(a)} Hazard Clearance Compliance (HCC) quantifies whether the trajectory maintains the prescribed clearance margin. \textbf{(b)} Hazard Proximity Response (HPR) characterizes whether the avoidance response is close to the minimum sufficient clearance.}
    \label{fig:eval}
    \vspace{-4pt}
\end{figure}

After a safety-critical hazard is inserted into a real driving scene, the original human trajectory may pass through the newly introduced hazard and therefore cannot serve as a valid reference trajectory for the edited scene. 
We therefore evaluate whether the predicted trajectory satisfies the task-defined local safety constraints, rather than using an unique trajectory label.
Let the corresponding source and edited trajectories, $\tau_i^o$ and $\tau_i^e$, for scene $i$ be: 


\begin{equation}
    \tau_i^o = \left\{(x_{i,t}^o, y_{i,t}^o)\right\}_{t=0}^{T},\tau_i^e = \left\{(x_{i,t}^e, y_{i,t}^e)\right\}_{t=0}^{T}.
    \label{eq:trajectory}
\end{equation}
The source prediction $\tau_i^o$ is used only as a counterfactual reference for diagnosing behavioral changes, and the safety of $\tau_i^e$ is evaluated by its geometric relationship with the task-defined hazard region $H_i$.

\emph{Counterfactual Trajectory Shift} (CTS) measures how strongly the planner changes its predicted trajectory after the hazard is introduced:
\begin{equation}
    {CTS} =\frac{1}{N}\sum_{i=1}^{N} \frac{1}{T+1}\sum_{t=0}^{T}\sqrt{\left(x_{i,t}^e-x_{i,t}^o\right)^2 + \left(y_{i,t}^e-y_{i,t}^o\right)^2}.
    \label{eq:cts}
\end{equation}
and its longitudinal and lateral components are given in the supplementary material. CTS diagnoses the planner's sensitivity to the counterfactual hazard, but does not by itself determine whether the resulting response is safe or directionally appropriate.

\emph{Unsafe Rate} (UR) measures direct intrusion into the task-defined core hazard region. 
We construct an ego collision tube $\mathbf{coll}_i$ from the edited trajectory using oriented vehicle footprints, with details given in the supplementary material.
Let $H_i$ denote the point set of the hazard region defined for scene $i$. 
We define the unsafe-region intrusion indicator as: 
\begin{equation}
    {ur_i} = \mathbb{I}\left[\mathbf{coll}_i \cap H_i \neq \varnothing \right],
    \label{eq:URi}
\end{equation}
where $\mathbb{I}[\cdot]$ is the indicator function, which equals one when the enclosed condition holds and zero otherwise. The dataset-level Unsafe Rate is: 
\begin{equation}
    {UR} =     \frac{1}{N}\sum_{i=1}^{N} ur_i .
    \label{eq:UR}
\end{equation}
A lower UR indicates that the planner enters the localized hazard or no-go region in fewer edited cases.


Avoiding direct intrusion is necessary but insufficient, because a safe trajectory may still pass unacceptably close to the hazard. Therefore, \emph{Hazard Clearance Compliance} (HCC) is used as our primary metric to measure geometric safety-compliance . 
Let $d_i =d(\mathbf{coll}_i ,H_i), d_i\ge 0$ denote the minimum Euclidean distance between the ego collision tube and the hazard region. 
Given the required safety clearance $d_{\mathrm{safe}}$, HCC is defined as: 
\begin{equation}
    {HCC}=\frac{1}{N}\sum_{i=1}^{N}(1-ur_i)\cdot\min\left(\frac{d_i}{d_{\mathrm{safe}}}, 1\right).
    \label{eq:hcc}
\end{equation}
HCC is bounded in $[0,1]$. It is zero when the trajectory intrudes into the hazard region, increases with clearance for non-overlapping trajectories, and saturates once the required safety clearance is reached.

\emph{Hazard Proximity Response} (HPR): although a trajectory should maintain sufficient clearance, its distance from the hazard also provides information about the magnitude of the avoidance response. 
We define HPR as: 
\begin{equation}
{HPR}=\frac{1}{N}\sum_{i=1}^{N}(1-ur_i)\cdot\frac{d_{\mathrm{safe}}}{\left|d_i-d_{\mathrm{safe}}\right|+d_{\mathrm{safe}}}.
\label{eq:hpr}
\end{equation}
HPR is zero when the trajectory enters the hazard region. 
Among safety-compliant trajectories, it reaches one when the minimum clearance equals $d_{\mathrm{safe}}$ and decreases as the trajectory deviates farther from this clearance. 
Thus, HPR is used to diagnose whether safe avoidance is close to the minimum sufficient response.
In particular, a high HPR does not by itself imply safety compliance, because trajectories with $d_i<d_{\mathrm{safe}}$ may still obtain a positive HPR value; their insufficient clearance is explicitly captured by HCC.


Illustrated in Fig.~\ref{fig:eval}, CTS, UR, HCC, and HPR together characterize counterfactual sensitivity, direct hazard-region violation, spatial-clearance compliance, and clearance proportionality, respectively. 
Safety comparisons primarily rely on high HCC and low UR, whereas CTS and HPR provide complementary diagnostic information and should not be interpreted independently as monotonic safety scores.

\subsection{Task-Grounded Risk Reminder}
\label{subsec:reminder_method}
The planner evaluation described above diagnoses whether a predicted trajectory responds safely to a localized counterfactual hazard, but it does not directly provide an interpretable explanation of the detected risk.  We therefore explore a secondary use of the ExceptionDrive taxonomy by constructing a lightweight task-grounded risk Reminder Agent for VLM-based driving agents. The Reminder Agent neither predicts trajectories nor controls the vehicle. Instead, it converts task-relevant visual evidence into a structured risk record that supports high-level driving strategy selection.

Given a driving scene, the reminder first identifies potential hazard evidence and associates it with the applicable risk definitions in the task taxonomy. 
It then produces a structured record containing  hazard presence, hazard type, and a recommended high-level strategy. 
The reminder has access to the task taxonomy as a collection of risk definitions, but is not provided with the ground-truth task identity or its target strategy for the current sample. 
The final driving strategy is predicted by a VLM-based driving agent from a predefined strategy space, including actions such as maintaining the current path, slowing down, stopping, yielding, or preparing to bypass. 


\section{Experiments}

\begin{figure}[!htbp]
\centering
\includegraphics[width=0.88\textwidth]{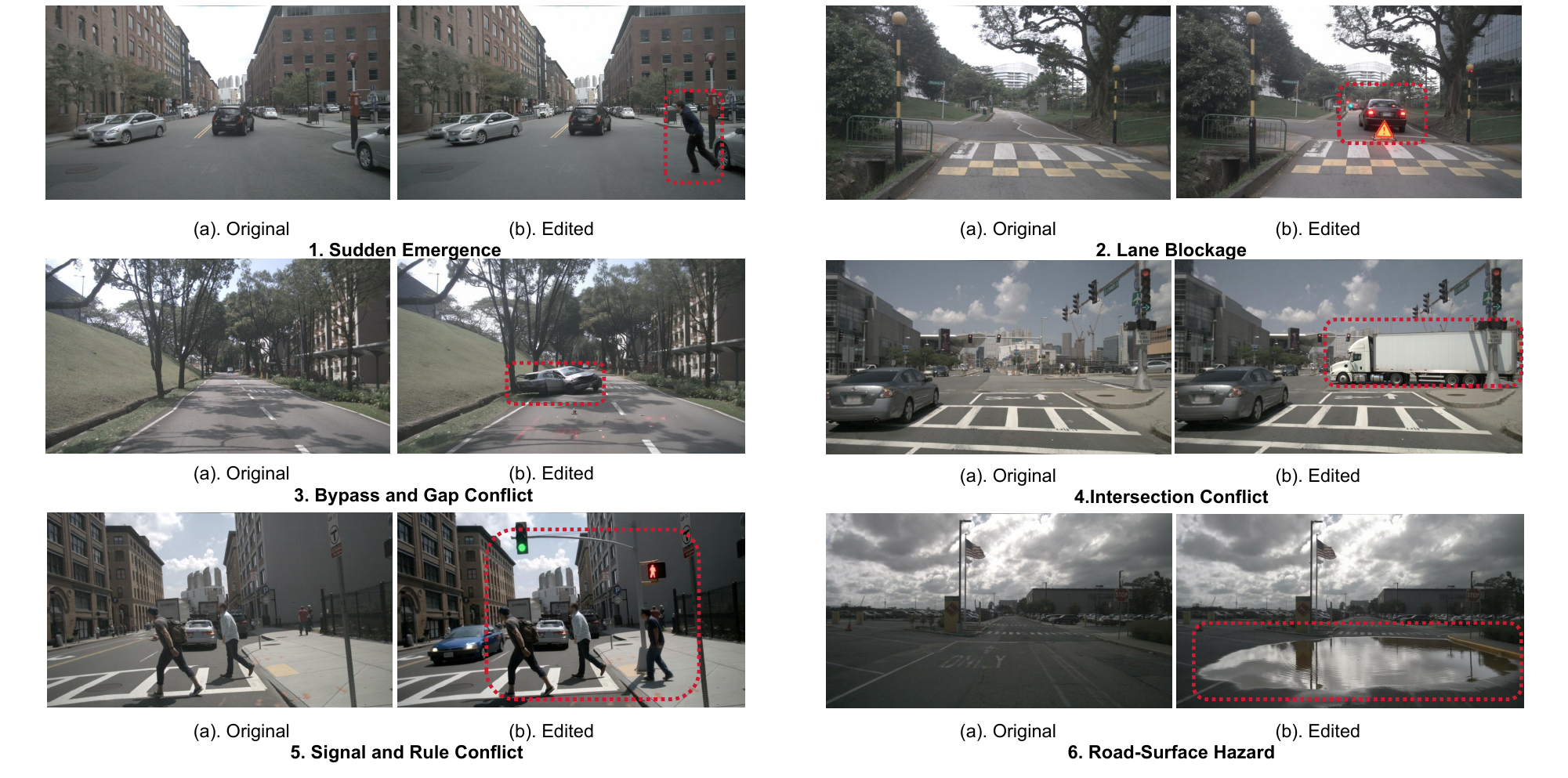}
\captionsetup{font=small,skip=3pt}
\caption{Representative samples from \textbf{ExceptionDrive} across six safety families. Each example pairs an original image with its counterfactual edit. Red dashed boxes mark the inserted local hazards. The edits preserve the original viewpoint, road layout, illumination, weather, and surrounding traffic context while modifying only the task-relevant risk factor.}
\label{dataset_samples}
\vspace{-4pt}
\end{figure}


\subsection{Dataset Statistics and Quality Analysis}

Starting from 850 nuScenes scenes and 21 predefined tasks, we obtain 17,850 scene–task combinations. Scene applicability filtering identifies 3,446 suitable pairs, accounting for 19.31\% of all combinations. After localized editing and VLM-based quality review, 3,181 samples are retained, corresponding to an acceptance rate of 92.31\%, and form the final benchmark.

To assess the reliability of the automated review, we randomly inspect 168 VLM-approved samples, covering 5.28\% of the benchmark. Among them, 163 pass manual inspection, yielding a 97.02\% acceptance rate and indicating a high manual confirmation rate among the audited VLM-approved samples.

Fig.~\ref{first_img} shows the distribution across six safety families. Sudden Emergence, Lane Blockage, Bypass and Gap Conflict, Signal and Rule Conflict, Intersection Conflict, and Road-Surface Hazard account for 25.56\%, 19.75\%, 24.96\%, 3.60\%, 9.10\%, and 17.03\%, respectively. Signal and Rule Conflict is the least frequent because it requires specific intersection layouts, traffic-light states, and right-of-way contexts.

To support failure analysis, qualitative studies, and future expansion, we construct an auxiliary set in addition to the benchmark evaluation set. Following the task distribution of the final benchmark, we select 8,994 additional scene--task pairs from outside the 3,446 screened benchmark candidates and process them directly with the editing model. These pairs are combined with the 265 edits rejected during benchmark quality review, yielding an auxiliary set of 9,259 pairs. The complete ExceptionDrive collection therefore contains \textbf{12,440} source--edited pairs: \textbf{3,181} quality-approved pairs in the benchmark evaluation set and \textbf{9,259} pairs in the auxiliary set. The auxiliary set is not used for the reported benchmark evaluation.

Fig.~\ref{dataset_samples} shows representative original–edited pairs. The approved samples introduce localized task-relevant hazards while largely preserving irrelevant scene content, enabling controlled counterfactual evaluation.

\subsection{Planner Evaluation on Corner Cases}

We evaluate seven representative planners on the final benchmark: Impromptu VLA, OmniDrive, AutoVLA, DrivoR, LightEMMA, ST-P3, and VAD. For each original–edited pair, the navigation command, historical information, and inference configuration remain unchanged. Model-specific input modalities and adaptations to the edited views are described in the supplementary material.

\begin{table}[t]
\centering
\scriptsize
\setlength{\tabcolsep}{3.2pt}

\resizebox{0.48\textwidth}{!}{
\begin{tabular}{l*{6}{r}}
\toprule
Model & HCC\% $\uparrow$ & UR\% $\downarrow$ & HPR\% $\uparrow$ & CTS & CTS$_x$ & CTS$_y$ \\
\midrule
Impromptu VLA~\cite{impromptu} &19.64 &78.92  &\textbf{18.87} &2.20 &1.86 &0.70 \\
OmniDrive~\cite{omnidrive}    &9.52  &83.36  &9.84 &1.40   &1.34   &0.22   \\
AutoVLA~\cite{autovla}       &10.37 & 81.93  &11.26 &1.97 &1.67 &0.64 \\
DrivoR~\cite{drivoR}      &4.86 &92.13 &5.42 &1.25 &1.19 &0.22 \\
LightEMMA~\cite{lightemma}    &18.66 &79.51  &18.04 &1.68 &1.61 &0.23 \\
ST-P3~\cite{stp3}    &\textbf{21.03} &\textbf{78.11}  &18.59 &1.57 &1.55 &0.24 \\
VAD~\cite{vad}    &13.89 &80.26  &13.47 &1.68 &1.47 &0.51 \\
\bottomrule
\end{tabular}%

}

\caption{Main planner evaluation results on the proposed corner-case benchmark. HCC is the primary metric and measures whether the planner produces the task-defined safe response in edited hazardous scenes. UR measures unsafe response into the annotated hazard region. HPR measures the efficiency of hazard avoidance. CTS reports the magnitude of counterfactual trajectory shift between source and edited inputs, with longitudinal and lateral components denoted by CTS$_x$ and CTS$_y$.}
\label{tab:planner-benchmark}
\end{table}

Table~\ref{tab:planner-benchmark} shows limited local hazard compliance across all planners: the highest HCC is 21.03\%, and every UR exceeds 78\%. ST-P3 performs best, with the highest HCC (21.03\%) and lowest UR (78.11\%), whereas DrivoR obtains the lowest HCC (4.86\%) and highest UR (92.13\%). Impromptu VLA produces the largest CTS (2.20) without achieving the best safety performance, confirming that trajectory change indicates sensitivity to the edit but not necessarily a safe response. We therefore use HCC and UR as the primary safety metrics and treat HPR and CTS as complementary diagnostics.


\begin{table}[t]
\centering
\small

\resizebox{0.47\textwidth}{!}{
\begin{tabular}{lcccccc}
\toprule
Planner & Emergence & Blockage & Bypass & Signal & Inter. & Surface \\
\midrule
autoVLA & 9.90 & 12.86 & 10.22 & 9.78 & 9.92 & 9.56 \\
Impromptu VLA & 19.63 & \textbf{20.28} & \textbf{19.98} & 20.14 & 17.86 & 19.95 \\
DrivoR &4.31 & 4.59 & 2.61 & 5.49 & 3.94 & 8.27 \\
ST-P3 &\textbf{20.02} & 19.86 & 17.48 & \textbf{22.92} & \textbf{24.75} & \textbf{21.17} \\
\bottomrule

\end{tabular}
}

\caption{Family-level HCC. The table reports planner response correctness across the six safety families, revealing which types of corner cases are most challenging for each model. Higher HCC indicates a larger fraction of edited hazardous scenes in which the planner produces the task-defined safe response.}
\label{tab:planner-family}
\end{table}

Table~\ref{tab:planner-family} shows that ST-P3 achieves the highest HCC in most families, while DrivoR reaches only 2.61\% on Bypass and Gap Conflict. For Signal and Rule Conflict and Intersection Conflict, HCC measures spatial compliance with task-defined regions rather than complete traffic-rule understanding.

\subsection{Qualitative Analysis}
\begin{figure}[!htbp]
\centering
\includegraphics[width=0.95\textwidth]{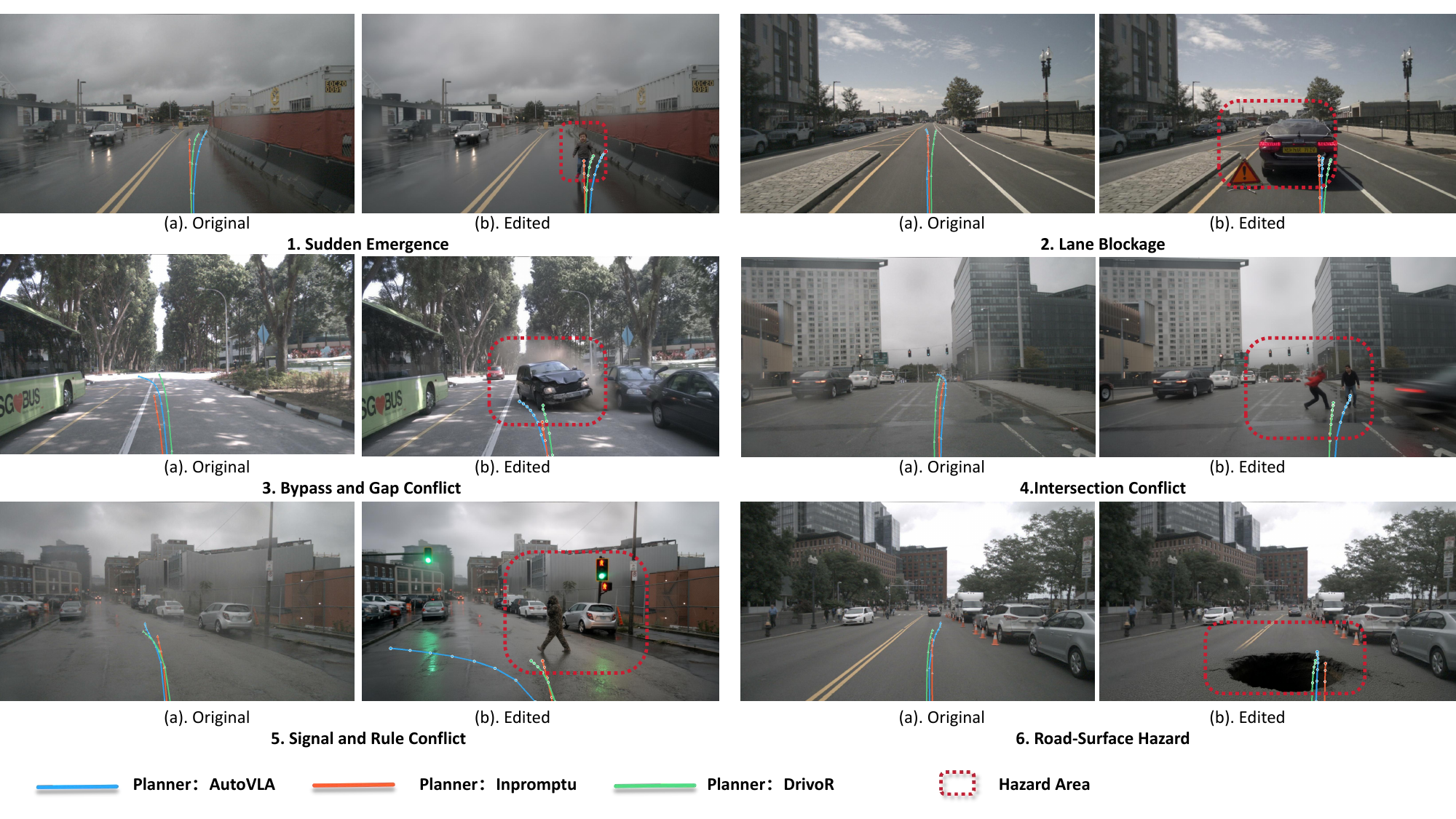}
\caption{Qualitative planner responses across the six \textbf{ExceptionDrive} safety families. For each family, (a) shows the source scene and (b) the edited counterfactual scene. Red dashed boxes denote the inserted hazards, while blue, orange, and green curves show trajectories from AutoVLA, Impromptu VLA, and DrivoR, respectively.}
\label{planner_case}
\end{figure}
Fig.~\ref{planner_case} presents representative planning cases and illustrates three common failure patterns. First, some models follow trajectories close to those predicted in the original scene and directly enter the risk region, resulting in increasing UR and decreaseing HCC. Second, some models react to the edit but shift in an inappropriate direction or by an insufficient amount, producing a large CTS without achieving safe avoidance. Third, some trajectories avoid direct intrusion but remain too close to the risk boundary, leading to a relatively lower UR but limited HCC.

These cases show that intrusion, insufficient clearance, and counterfactual trajectory change are related but non-equivalent phenomena. The qualitative results therefore complement, rather than replace, the quantitative metrics.

\subsection{Reminder Agent Evaluation}

We further evaluate task-grounded reminders as a downstream application of ExceptionDrive, using Qwen2.5-VL-7B~\cite{qwenVL} as the final driving agent. The baseline and reminder-augmented settings use identical visual inputs, candidate strategies, and base instructions. The baseline predicts directly from the scene, whereas the augmented setting additionally receives the structured reminder defined in Section~\ref{subsec:reminder_method}, generated by an auxiliary VLM from the same visual input and shared task taxonomy. Neither setting receives human-annotated risk regions, sample-specific safe actions, or quality-review labels.

To avoid ambiguity in free-form evaluation, Qwen2.5-VL-7B selects one of five predefined strategies, denoted A--E. A prediction is correct if it belongs to the task-defined acceptable response set. We report strategy accuracy and under warning rate, where the latter measures the fraction of hazardous cases in which the selected action is insufficiently cautious.

\begin{table}[h]
\centering
\small
\setlength{\tabcolsep}{3.5pt}

\resizebox{0.45 \textwidth}{!}{
\begin{tabular}{lccc}
\toprule
Method & Accuracy $\uparrow$ & Under-warning Rate $\downarrow$ \\
\midrule
Qwen2.5VL-7B & 51.86 & 45.53 \\
Qwen2.5VL-7B + Reminder & 70.71 & 29.18 \\
\bottomrule
\end{tabular}

}
\caption{High-level strategy evaluation for the zero-shot VLM driving agent.}
\label{tab:VLM+reminder}
\end{table}

Table~\ref{tab:VLM+reminder} shows that the reminder improves strategy accuracy from 51.86\% to 70.71\% and reduces the miss rate from 45.53\% to 29.18\%. These results demonstrate the value of ExceptionDrive for both planner diagnosis and structured risk communication to VLM-based driving agents.

\section{Conclusion}

In this paper, we proposed \textbf{ExceptionDrive}, a planning oriented counterfactual benchmark for evaluating autonomous-driving planners under rare and safety-critical local hazards. ExceptionDrive constructs 21 tasks across six planning-safety families by inserting localized hazards into compatible real-world driving observations while preserving the original scene context. To address the absence of valid expert trajectories after hazard insertion, we proposed a hazard-centric evaluation protocol that assesses hazard intrusion, clearance compliance, avoidance response, and counterfactual behavioral changes. In addition, we developed a lightweight Reminder Agent to transform visual hazard evidence into structured risk records for high-level decision making. Experiments with representative planners demonstrate that existing systems frequently fail to maintain sufficient safety clearance, while the Reminder Agent improves zero-shot strategy prediction and reduces under-warning. These results highlight the value of ExceptionDrive for systematically analyzing planner behavior under localized visual interventions that are critical to safety.


\bibliographystyle{plainnat}
\bibliography{references}

\end{document}